\documentclass[preprint,3p]{elsarticle}

\journal{arXiv}
\usepackage[utf8]{inputenc}
\usepackage{bm}
\usepackage{amsmath,comment}
\usepackage[hidelinks]{hyperref}
\hypersetup{colorlinks=true,linkcolor=blue,urlcolor=blue}
\usepackage{multicol}
\usepackage{booktabs}
\usepackage{amsthm}
\newtheorem{remark}{Remark}
\usepackage[table,xcdraw,dvipsnames]{xcolor}
\usepackage{graphicx}
\usepackage{float}
\usepackage{subcaption}
\usepackage{algorithm}
\usepackage{algpseudocode}
\usepackage{pdflscape}
\usepackage{array}
\usepackage{multirow}
\usepackage{diagbox}
\usepackage{cleveref}
\usepackage{amsfonts}
\usepackage{amssymb}
\usepackage{ulem}
\usepackage{fourier-orns}
\usepackage[labelfont=bf,labelsep=period]{caption}
\usepackage[labelfont=rm,font=footnotesize]{subcaption}

\numberwithin{equation}{section}

\DeclareMathAlphabet{\altmathcal}{OMS}{cmsy}{m}{n}
\DeclareMathAlphabet{\altmathcalb}{OMS}{cmsy}{b}{n}
\DeclareMathAlphabet{\mathcalboondox}{U}{BOONDOX-calo}{m}{n}
\DeclareMathAlphabet{\mathbbmsl}{U}{bbm}{m}{sl}

\newcommand{\orcid}[1]{\href{https://orcid.org/#1}{\texorpdfstring{\includegraphics*[width=8pt]{figs/orcid}}~~}{}}

\definecolor{drot}{rgb}{0.7,0,0.1}

\begin{document}
\baselineskip14pt
\sloppy

\begin{frontmatter}

\title{RL-DAUNCE: Reinforcement Learning-Driven Data Assimilation with Uncertainty-Aware Constrained Ensembles}

\author[UW]{Pouria Behnoudfar}
\author[UW]{Nan Chen\corref{corr}}
\cortext[corr]{Corresponding author}

\address[UW]{{University of Wisconsin-Madison, Department of Mathematics, Madison, WI, USA}}

\begin{abstract}
Machine learning has become a powerful tool for enhancing data assimilation. While supervised learning remains the standard method, reinforcement learning (RL) offers unique advantages through its sequential decision-making framework, which naturally fits the iterative nature of data assimilation by dynamically balancing model forecasts with observations. In this paper, we develop RL-DAUNCE (Reinforcement Learning-Driven Data Assimilation with Uncertainty-Aware Constrained Ensembles), a new RL-based method that enhances data assimilation with physical constraints through three key aspects. First, RL-DAUNCE naturally inherits the computational efficiency of machine learning while it uniquely structures its agents to mirror ensemble members in conventional data assimilation methods. Second, RL-DAUNCE emphasizes uncertainty quantification by advancing multiple ensemble members, moving beyond simple mean-state optimization. Third, RL-DAUNCE's ensemble-as-agents design inherently facilitates the enforcement of physical constraints during the assimilation process, which is crucial to improving the state estimation and subsequent forecasting. A primal-dual optimization strategy is developed to enforce constraints, which dynamically penalizes the reward function to ensure constraint satisfaction throughout the learning process. In addition, state variable bounds are respected by constraining the RL action space. Together, these features ensure physical consistency without sacrificing efficiency. RL-DAUNCE is applied to the Madden–Julian Oscillation, an intermittent atmospheric phenomenon characterized by strongly non-Gaussian features and multiple physical constraints. RL-DAUNCE outperforms the standard ensemble Kalman filter (EnKF), which fails catastrophically due to the violation of physical constraints. Notably, RL-DAUNCE matches the performance of constrained EnKF, particularly in recovering intermittent signals, capturing extreme events, and quantifying uncertainties, while requiring substantially less computational effort.
\end{abstract}
\end{frontmatter}
\section{Introduction}






Data assimilation is a systematic technique used to refine the state estimate of a physical system by combining observational data with prior model information~\cite{law2015data, majda2012filtering, asch2016data, evensen2022data}. Originally developed for numerical weather prediction to improve forecast initialization, data assimilation now has broad applications, including parameter estimation, dynamical interpolation, control assistance, and model identification~\cite{Freitag2013Synergy, chen2023stochastic, wang2000data}. More recently, data assimilation has also become a key component in digital twins \cite{zhong2023reduced, donato2024self, chen2025cgkn} and multi-model forecasting systems \cite{bach2023multi, chen2019multi, stephenson2005forecast}.

The classical Kalman filter \cite{kalman1960new} provides an optimal state estimate for linear and Gaussian systems through closed-form analytical solutions. However, in strongly nonlinear and non-Gaussian scenarios, approximate numerical methods, such as the extended Kalman filter, variational data assimilation, and ensemble or particle-based techniques, are necessary to handle these practical issues~\cite{ribeiro2004kalman, keppenne2002initial, Carrassi2008Data, evensen1994sequential, chen2016filtering, merwe2004sigma}. Among these, the ensemble Kalman filter (EnKF) and its variants stand out as some of the most widely adopted approaches \cite{evensen2003ensemble, houtekamer2005ensemble}. EnKF has been successfully applied across diverse fields, including meteorology~\cite{Anderson1996Data, kalnay2003atmospheric}, soil and land studies~\cite{lievens2016assimilation, de2012multiscale, Li2024Land}, groundwater modeling~\cite{vrugt2003shuffled}, and broader geophysical applications, such as integrating Global Observing System data~\cite{lahoz2014data} and remote sensing observations~\cite{li2020harmonizing}.
Despite the widespread use of the EnKF and other numerical methods, they face several common computational challenges. The high dimensionality of these problems often slows down calculations, particularly because the methods require repeated runs of the forecast model to generate ensembles~\cite{Dong2023hybrid}. In practice, only a limited number of ensembles are available. Therefore, additional empirical tuning, such as noise inflation and localization~\cite{anderson2012localization, anderson2007exploring, buehner2007spectral, whitaker2012evaluating}, is frequently necessary to ensure the numerical stability of these data assimilation methods.

Recently, machine learning has been widely adopted to enhance data assimilation~\cite{cheng2023machine}. Existing machine learning approaches for data assimilation can be broadly categorized into two groups: (1) end-to-end machine learning for data assimilation~\cite{revach2022kalmannet, boudier2020dan, ouala2018neural, manucharyan2021deep, mou2023combining, boudier2023data} and (2) replacing traditional forecast models with machine learning-based forecast models~\cite{gottwald2021combining, chattopadhyay2023deep, gilbert2010machine, wang2016auto, otto2019linearly, takeishi2017learning, brajard2020combining, gottwald2021supervised, buizza2022data, farchi2021using, bocquet2020bayesian, tsuyuki2022nonlinear, maulik2022aieada, penny2022integrating, pawar2020long, tang2020deep}. These methods have significantly improved data assimilation efficiency and demonstrated promising results. Yet, there is still great potential for machine learning to enhance the effectiveness of data assimilation, particularly in achieving more physically consistent state estimates in the presence of noisy observations. Physical consistency includes not only adhering to physical laws and constraints but also advancing robust uncertainty quantification \cite{janjic2014conservation, chattopadhyay2021towards, majda2018model}.

Reinforcement Learning (RL) is a machine learning paradigm in which an agent learns optimal decision-making policies by interacting with an environment to maximize cumulative rewards \cite{kaelbling1996reinforcement, wiering2012reinforcement}. Unlike supervised learning, which is the standard machine learning approach for data assimilation, RL specializes in sequential decision-making through trial-and-error optimization, which naturally fits the iterative nature of data assimilation by dynamically balancing model forecasts with observations. By framing the data assimilation cycle as an environment and assimilation accuracy as a reward signal, RL can autonomously develop adaptive strategies that outperform static algorithms, especially in non-stationary or chaotic systems where traditional methods struggle. Recent advances in deep RL have further expanded the applicability of RL to high-dimensional state spaces through neural network function approximators \cite{raffin2021stable, feng2023deep}, which allows RL to be applicable to practical data assimilation scenarios.


In this paper, we develop RL-DAUNCE (Reinforcement Learning-Driven Data Assimilation with Uncertainty-Aware Constrained Ensembles), a new RL-based method that enhances data assimilation through three key aspects. First, RL-DAUNCE naturally inherits the computational efficiency of machine learning while it uniquely structures its agents to mirror ensemble members in conventional DA methods. It maintains compatibility with standard workflows while leveraging the adaptive capabilities of RL. Second, unlike existing RL-assisted DA methods focusing on end-to-end state estimation~\cite{hammoud2024data}, RL-DAUNCE emphasizes uncertainty quantification by advancing multiple ensemble members, moving beyond simple mean-state optimization. It aims to provide an uncertainty-aware state estimation. Third, RL-DAUNCE's ensemble-as-agents design inherently facilitates the enforcement of physical constraints during assimilation. This is achieved by incorporating a primal-dual optimization strategy into the RL to enforce various constraints, which dynamically penalizes the reward function to ensure constraint satisfaction throughout the learning process. In addition, state variable bounds are respected by constraining the RL action space. Together, these features ensure physical consistency without sacrificing efficiency. The constrained data assimilation has been shown to provide physically consistent solutions that outperform standard EnKF in both accuracy and robustness \cite{janjic2021weakly, ruckstuhl2021training, janjic2014conservation}. Importantly, constrained data assimilation not only improves state estimation but also yields more reliable forecasts and results in higher-quality reanalysis datasets \cite{gleiter2022ensemble}. These features demonstrate RL-DAUNCE's dual advantage in both computational efficiency and physical fidelity.

The remainder of this paper is organized as follows. Section~\ref{sec:KF} provides an overview of the EnKF and its constrained formulation. Section~\ref{sec:RK_DA} discusses the conceptual parallels between RL and the EnKF, followed by the development of the RL-DAUNCE framework. Section~\ref{sec:MJO} demonstrates RL-DAUNCE's application to the Madden–Julian Oscillation (MJO), an intermittent atmospheric phenomenon characterized by strongly non-Gaussian features and multiple physical constraints. Concluding remarks are presented in Section~\ref{sec:conclude}.

\section{Overview of the Ensemble Kalman Filter (EnKF) and its Constrained Formulation}
\label{sec:KF}
This section provides an overview of EnKF and its constrained formulation. They will be used as state-of-the-art methods to compare with the RL-DAUNCE, and the constrained EnKF will also provide training data for the RL-DAUNCE.
\subsection{Ensemble Kalman filter (EnKF)}

The EnKF generalizes the standard Kalman filter to nonlinear systems by approximating the state distribution through a finite ensemble of possible state realizations~\cite{evensen2003ensemble, houtekamer2005ensemble, burgers1998analysis}. Unlike the classical Kalman filter, which propagates analytical mean and covariance estimates, the EnKF instead propagates an ensemble of $N$ state vectors through a two-step forecast-analysis procedure.

In the forecast step, the EnKF utilizes the forecast model to update the background ensemble $\{\mathbf{x}_k^{b,(i)}\}_{i=1}^N$ with perturbed observations $\{\mathbf{z}_k^{(i)}\}_{i=1}^N$. Specifically, each ensemble member is propagated forward in time using the system dynamics:
\begin{equation}
    \mathbf{x}_k^{b, (i)} = \mathbf{A}_{k-1}\left(\mathbf{x}_{k-1}^{b,(i)}\right) + \boldsymbol\sigma_{k-1}^{(i)}, \quad i = 1, 2, \dots, N,
    \label{eq:enkf_forecast}
\end{equation}
where $\mathbf{A}_{k-1}$ is a nonlinear operator and $\boldsymbol\sigma_{k-1}^{(i)} \sim \mathcal{N}(\mathbf{0}, \mathbf{Q})$ represents process noise sampled for each ensemble member. The sample mean and covariance of the background ensembles are computed as:
\begin{equation}
    \bar{\mathbf{x}}_k = \frac{1}{N} \sum_{i=1}^{N} \mathbf{x}_k^{b,(i)}, \quad \mathbf{P}_k = \frac{1}{N - 1} \sum_{i=1}^{N} (\mathbf{x}_k^{b,(i)} - \bar{\mathbf{x}}_k)(\mathbf{x}_k^{b,(i)} - \bar{\mathbf{x}}_k)^T.
    \label{eq:enkf_mean_covariance}
\end{equation}

Then in the analysis (update) step, given an observation $\mathbf{z}_k$, the measurement equation for each ensemble member is:
\begin{equation}
    \mathbf{z}_k^{(i)} = \mathbf{H}_k \left(\mathbf{x}_k^{(i)}\right) + \boldsymbol\sigma_{o,k}^{(i)}, \quad i = 1, 2, \dots, N,
    \label{eq:enkf_measurement}
\end{equation}
where the observation is perturbed as $\mathbf{z}_k^{(i)}  = \mathbf{z}_k  + \boldsymbol\varepsilon_{k}^{(i)}$, with $\boldsymbol\sigma_{o,k}^{(i)}, \boldsymbol\varepsilon_{k}^{(i)} \sim \mathcal{N}(\mathbf{0}, \mathbf{R})$.
The Kalman gain is then computed as:
\begin{equation}
    \mathbf{K}_k = \mathbf{P}_k \mathbf{H}^T(\mathbf{H} \mathbf{P}_k \mathbf{H}^T + \mathbf{R})^{-1},
    \label{eq:enkf_kalman_gain}
\end{equation}
where $\mathbf{H}$ denotes the linear observation operator. Each ensemble member is updated using~\cite{evensen2003ensemble}:
\begin{equation}
\label{eq:EnKF}
    \mathbf{x}_k^{a,(i)} = \mathbf{x}_k^{b,(i)} + \mathbf{K}_k \left(\mathbf{z}_k^{(i)}  - \mathbf{H} \left(\mathbf{x}_k^{b,(i)} \right)\right), \quad i = 1, 2, \dots, N.
\end{equation}
There are many practical strategies to improve the robustness of the EnKF and mitigate sampling errors, such as the ensemble transform Kalman filter (ETKF) \cite{bishop2001adaptive, hunt2007efficient} and the ensemble adjustment Kalman filter (EAKF) \cite{anderson2001ensemble} or in general the family of the ensemble square root filter (EnSRF) \cite{whitaker2002ensemble,thomas2009robust}.
Localization is also a technique widely incorporated into the ensemble-based data assimilation approaches for high-dimensional systems to mitigate the impact of sampling errors, as only a small number of samples is affordable in practice. It can effectively ameliorate the spurious long-range correlations between the background and the observations. Some practical localization approaches can be found in \cite{hunt2007efficient, anderson2012localization, anderson2007exploring, fertig2007assimilating, campbell2010vertical, houtekamer2016review}.
%
\subsection{Constrained EnKF}

The standard EnKF provides an efficient framework for state estimation by updating an ensemble of system states based on observational data. However, it does not inherently enforce physical constraints, such as positivity preservation or energy conservation, which are often critical in scientific and engineering applications. The EnKF update step can be reformulated to address this limitation as a constrained minimization problem~\cite{janjic2014conservation, gleiter2022ensemble}. By casting the assimilation process in an optimization framework, we can explicitly impose constraints and leverage numerical optimization techniques to obtain physically consistent updates while preserving the statistical properties of the EnKF.

The constrained EnKF solves the following optimization problem for each ensemble member to enforce physical constraints~\cite{janjic2014conservation,gleiter2022ensemble}:
\begin{equation}
    \mathbf{x}_k^{a,(i)} = \arg \min_{\mathbf{x}} J_k^{(i)}(\mathbf{x}),
\end{equation}
subject to constraints
\begin{equation}
    c_{eq}(\mathbf{x}) = 0, \quad \text{and/or} \quad c_{neq}(\mathbf{x}) \leq 0,
\end{equation}
with \( c_{eq} \) and \( c_{neq} \) representing (nonlinear) equality and inequality constraint functions. The minimization of the cost functions \( J_k^{(i)} \), subject to these constraints, yields the analysis ensemble members \( x_{k}^{a,(i)} \) at time step \( k \). Following the notations in the previous subsection, the cost function reads:
\begin{equation}
\label{eq:enkf}
    J_k^{(i)}(\mathbf{x}) = \frac{1}{2} \left( \mathbf{x} - \mathbf{x}_k^{b,(i)} \right)^\top \mathbf{P}_k^{-1} \left( \mathbf{x} - \mathbf{x}_k^{b,(i)} \right) + \frac{1}{2} \left( \mathbf{z}_k^{(i)} - \mathbf{H} (\mathbf{x}) \right)^\top \mathbf{R}^{-1} \left( \mathbf{z}_k^{(i)} -\mathbf{H} (\mathbf{x}) \right).
\end{equation}

\begin{remark}
   In the case of a linear observation operator, i.e., \( \mathbf{H}(x_k) = \mathbf{H_k} \), the cost functions take a quadratic form, allowing for a reformulation of the problem. This quadratic structure makes the Hessians and gradients of the objective functions explicitly identifiable \cite{gleiter2022ensemble}. Introducing a variable transformation as \( \mathbf{x}_{k}^{a,(i)} \rightarrow \mathbf{z}_{k}^{a,(i)} \), the problem reads:
\[
\mathbf{z}_{k}^{a,(i)} = \arg\min_{\mathbf{z}} {J}_k^{(i)} (\mathbf{z}),
\]
where the cost function is defined as
\[
J_k^{(i)} (\mathbf{z}) = \frac{1}{2} \mathbf{z}^\top \left( I + (\mathbf{H}_k \mathbf{X}_{k}^{L})^\top \mathbf{R}^{-1} \mathbf{H}_k \mathbf{X}_{k}^L \right) \mathbf{z}
+ \left( \mathbf{H}_k \mathbf{x}_{k}^{b,(i)} - \mathbf{z}_k \right)^\top \mathbf{R}^{-1} \mathbf{H}_k \mathbf{X}_{k}^L \mathbf{z}.
\]
Here, \( \mathbf{X}_{k}^L \) represents the square root of the localized background-error covariance matrix, obtained via Cholesky decomposition, i.e.,
\[
\mathbf{P}_{k} = \mathbf{X}_{k}^L (\mathbf{X}_{k}^L)^\top,
\]
results in \[
\mathbf{x}_{k}^{a,(i)} = \mathbf{x}_{k}^{b,(i)} + \mathbf{X}_{k}^L \mathbf{z}_{k}^{a,(i)}.
\]
Since this transformation is linear, it preserves the linear or nonlinear nature of the constraints. Depending on the type of constraints (linear/nonlinear, equality/inequality), different numerical optimization algorithms can be employed, each with distinct computational requirements and convergence properties, yielding either local or global minima.

\end{remark}


\section{RL-DAUNCE Framework}
\label{sec:RK_DA}

In this section, we present the RL-DAUNCE framework. RL-DAUNCE establishes each ensemble member as an independent RL agent, implemented through separate policy networks. These agents are trained on datasets generated by constrained EnKF simulations to learn the temporal evolution of their respective ensemble members. To ensure physical consistency, we introduce a dual optimization scheme featuring dynamically adjusted Lagrange multipliers, which automatically balances prediction accuracy with physical constraints.

\subsection{Reinforcement learning (RL)}
RL is a machine learning paradigm in which an agent learns to make decisions by interacting with an environment and receiving feedback in the form of rewards~\cite{kaelbling1996reinforcement, wiering2012reinforcement}. RL aims to determine an optimal policy that maximizes cumulative rewards over time. Unlike supervised learning, where labeled data is provided, RL operates through trial and error, balancing exploration (trying new actions) and exploitation (choosing actions that yield high rewards)~\cite{ladosz2022exploration}.

Mathematically, the RL problems are typically formulated as a Markov Decision Process (MDP), defined by a tuple $(\mathcal{S}, \mathcal{A}, P, R, \gamma)$, where $\mathcal{S} \subset \mathbb{R}^n$ denotes the observation (state) space, and $\mathcal{A} \subset \mathbb{R}^n$ is the action space, $P(s'|s,a)$ represents the transition probabilities, $R(s,a)$ is the reward function, and $\gamma \in [0,1]$ is the discount factor that determines the importance of future rewards~\cite{dad2019introduction}. The policy $\pi_\theta: \mathcal{S} \rightarrow \mathcal{A}$ is a parameterized mapping from observations to actions, and the goal is to learn $\pi_\theta$ that maximizes the expected cumulative reward:
\begin{equation}
\label{eq:agent}
    \max_{\theta} \ \mathbb{E}_{s \sim p(s), \ a = \pi_\theta(s)} \left[ R(s, a) \right].
\end{equation}
The environment then transitions to a new state $s'$, based on the transition probabilities $P(s'|s,a)$. The agent iteratively updates its policy to maximize long-term rewards using an algorithm (here, actor-critic approaches).

In addition to learning optimal decision-making strategies, RL offers intrinsic mechanisms for uncertainty quantification. Many RL algorithms naturally account for uncertainty in policy estimates, value functions, and environment transitions through their agent behavior (e.g., exploration) or by exploring Bayesian-based algorithms. This feature is essential when operating in partially observed or stochastic environments, making RL particularly compatible with data assimilation frameworks. Uncertainty-aware RL agents can make more robust predictions by integrating model confidence levels, which aligns closely with the probabilistic nature of state estimation in filtering methods.

\subsection{Conceptual parallels between the RL and the EnKF}
Table~\ref{tab:ppo_vs_enkf} presents a conceptual comparison between Proximal Policy Optimization (PPO), a widely used RL algorithm, and EnKF. While the two frameworks originate from different domains, PPO from RL and EnKF from statistical estimation, they share several underlying principles. For instance, both methods rely on sampling techniques: PPO samples actions from a learned policy distribution, while EnKF samples state realizations from an ensemble. Additionally, both incorporate uncertainty quantification: PPO through stochastic policies and exploration mechanisms, and EnKF via ensemble statistics. Despite their differing objectives, PPO aims to maximize long-term rewards, whereas EnKF minimizes estimation error; their update mechanisms (policy gradients vs. Kalman gain) serve a similar role in iteratively refining the agent's behavior or the state estimate. This parallel highlights the potential of RL methods to complement and enhance data assimilation frameworks, particularly in handling uncertainty and learning dynamic update rules.

\begin{table}[h!]
\centering
\begin{tabular}{|l|l|l|}
\hline
\textbf{Aspect} & \textbf{PPO} & \textbf{EnKF} \\ \hline
Domain & Reinforcement Learning & Data Assimilation \\ \hline
Objective & Maximize reward & Minimize state estimation error \\ \hline
Updates & Policy gradient & Kalman gain \\ \hline
Uncertainty Handling & Stochastic policies & Ensemble representation \\ \hline
Sampling & Actions from policy & States from ensemble \\ \hline
\end{tabular}
\caption{Comparison between PPO and EnKF}
\label{tab:ppo_vs_enkf}
\end{table}

\subsection{Defining RL agents consistent with the EnKF}

To design an RL agent that mirrors the functionality of the EnKF, we adopt an ensemble-based formulation. Specifically, the RL agent is modeled as an ensemble of \(N\) policy networks:
\[
\pi_{\theta^{(i)}}(a^{(i)}|s^{(i)}), \quad i = 1, \dots, N,
\]
where \(\theta^{(i)}\) denotes the parameters of the \(i\)-th policy and $s^{(i)}$ encodes the relevant features of the forecast ensemble \(i\). Next, we formulate the optimization problem as: 

\begin{equation}
\label{eq:min}
\begin{aligned}
&\min_{\theta^{(i)}} \quad && \mathbb{E}_{s^{(i)} \sim p(s^{(i)}), (s^{(i)},a^*) \sim \mathcal{D}^{(i)}} \left[ \left\| \pi_{\theta^{(i)}}(s^{(i)}) - a^* \right\|^2 \right],
\end{aligned}
\end{equation}
with $\mathcal{D}^{(i)}$ denoting the dataset corresponds to the $i$-th ensemble. 
Here, each policy acts as an independent sample from the policy distribution, effectively capturing the uncertainty in learning the system's dynamic behavior. This ensemble setup enables robust state estimation and draws a direct analogy to the ensemble members used in the EnKF.

Furthermore, we formulate the learning task~\eqref{eq:min} as a constrained optimization problem to incorporate physical constraints, such as energy conservation. That is:
\begin{equation}
\label{eq:min_const}
\begin{aligned}
&\min_{\theta^{(i)}} \quad && \mathbb{E}_{s^{(i)} \sim p(s^{(i)}), (s^{(i)},a^*) \sim \mathcal{D}^{(i)}} \left[ \left\| \pi_{\theta^{(i)}}(s^{(i)}) - a^* \right\|^2 \right],\\
&\text{subject to} \quad && \mathbb{E}_{s^{(i)} \sim p(s^{(i)})} \left[ \delta E\left( \pi_{\theta^{(i)}}(s^{(i)}) \right) \right] \leq \epsilon,
\end{aligned}
\end{equation}
where $\delta E( \pi_{\theta^{(i)}}(s^{(i)}))$ measures the violation of the physical constraint (e.g., deviation from conserved energy), and $\epsilon > 0$ is a tolerance. 

Next, we provide a dataset generated using a constrained EnKF framework to train the RL agent. At each time step, the RL agent receives the current state variables of the \(i\)-th ensemble member, along with their temporal derivatives. This input structure allows the agent to learn not only from the current state but also from its immediate dynamics, mimicking the predictive step in the EnKF. The agent's action predicts the filtered state at the next time step.

The learning objective is to minimize the mean squared error (MSE) between the predicted action $a^{(i)}$ (i.e., the estimated state of ensemble $i$ at the next time step) and the reference solution at the next time step, provided by constrained EnKF. This formulation effectively trains the policy to emulate the filtering step of the EnKF while allowing for end-to-end learning of the update operator within an RL framework.

Using only the MSE between $a$ and the reference solution, an agent learns by assigning an instantaneous reward to each action taken in response to observations. These rewards are aggregated along trajectories to form the cumulative return, which measures the policy's performance. However, maximizing the cumulative reward at each episode cannot guarantee that the estimations are physically meaningful. Then, we explicitly enforce physical constraints during our RL training. To this end, we develop an algorithm based on the primal-dual optimization technique to enforce nonlinear conservative quantities (e.g., energy) and optimization over subspaces for direct constraints on state variables (e.g., positivity preserving).

\subsection{Primal-Dual optimization with dynamically adjusted Lagrange multiplier}
\label{sec:Dual_opt}

The primal-dual approach offers a robust and flexible framework for enforcing constraints in RL~\cite{paternain2022safe}. This method addresses the original RL problem (primal problem) of maximizing the expected reward while formally incorporating constraints through inequalities or equalities. Dual variables, or Lagrange multipliers, are introduced for each constraint, enabling the formulation of a Lagrangian function that combines the primal objective and the constraints.

The primal-dual approach is particularly effective in balancing the trade-off between exploration and exploitation while ensuring adherence to system constraints and quantifying uncertainty. It allows the RL model to learn the dynamical behavior of the system accurately without compromising on constraint satisfaction or physical consistency.

To enforce the linear or nonlinear constraint \(\delta E(a^{(i)}) \leq \epsilon \) of \eqref{eq:min_const} in RL-DAUNCE data assimilation framework, we use a Lagrange multiplier \( \lambda\left(s^{(i)}\right) \) to penalize constraint violations dynamically based on the provided observations. The Lagrange multiplier is updated based on the severity of the violation at each marching step, ensuring effective constraint enforcement while preventing \( \lambda\left(s^{(i)}\right) \) from growing unbounded. Therefore, we define the Lagrangian as:
\begin{equation}
\label{eq:lagrange}
  \mathcal{L}\left(\theta^{(i)}, s^{(i)}\right) = \mathbb{E}_{s^{(i)} \sim p(s^{(i)}), (s^{(i)},a^*) \sim \mathcal{D}^{(i)}} \left[ \left\| \pi_{\theta^{(i)}}(s^{(i)}) - a^* \right\|^2 \right]+  \lambda\left(s^{(i)}\right) \left( \mathbb{E}_{s^{(i)} \sim p(s^{(i)})} \left[ \delta E\left( \pi_{\theta^{(i)}}(s^{(i)}) \right) \right] - \epsilon \right).
\end{equation}
Given the reward function $R(s^{(i)},a^{(i)})$, \eqref{eq:lagrange} is explained as
\begin{equation}
\label{eq:lag}
     \mathcal{L}\left(\theta^{(i)}, s^{(i)}\right) = R\left(s^{(i)},a^{(i)}\right) - \lambda\left(s^{(i)}\right)  \cdot \left(\delta E(a^{(i)}) - \epsilon\right),
\end{equation}
where   $\mathcal{L}\left(\theta^{(i)}, s^{(i)}\right)$ is the Lagrangian defined at each iteration step for the ensemble $i$. This dependency is expressed as a function of observation $s^{(i)}$. $\lambda\left(s^{(i)}\right) \geq 0$ is the dual variable associated with the constraint. Accordingly, the agent's goal becomes solving the min-max problem:
\[
\max_{\theta^{(i)}} \min_{\lambda \geq 0}  \mathcal{L}\left(\theta^{(i)}, s^{(i)}\right).
\]

Finally, we can write our algorithm structure as two steps of


\begin{enumerate}
    \item \textbf{Policy (Primal) Step:}  
    Update the policy $\pi_{\theta^{(i)}}$ by performing (approximate) gradient ascent on the Lagrangian $\mathcal{L}\left(\theta^{(i)}, s^{(i)}\right)$:
    \begin{equation}
        \theta^{(i)} \leftarrow \theta^{(i)} - \alpha_{\theta^{(i)}} \nabla_{\theta^{(i)}} \mathcal{L}\left(\theta^{(i)}, s^{(i)}\right).
    \end{equation}

    \item \textbf{Dual Step:}  
    Update the Lagrange multiplier $\lambda$ by performing gradient ascent on the dual objective:
    \begin{equation}
    \label{eq:dual_expect}
\lambda\left(s^{(i)}\right)\leftarrow \left[ \lambda\left(s^{(i)}\right) + \alpha_\lambda \left( \mathbb{E}_{s^{(i)} \sim p(s^{(i)})} \left[ \delta E\left( \pi_{\theta^{(i)}}\left(s^{(i)}\right) \right) \right] - \epsilon \right) \right]_+,
    \end{equation}
    projecting onto the non-negative orthant to ensure $\lambda\left(s^{(i)}\right) \geq 0$. Since there is no guarantee that the dual problem~\eqref{eq:dual_expect} converges, we write the constraint as $\frac{1}{\delta E} > \frac{1}{\epsilon}$. Then, the multiplier is updated as:
    \begin{equation}
    \label{eq:dual_ex}
\lambda\left(s^{(i)}\right)\leftarrow \left[ \lambda\left(s^{(i)}\right) - \alpha_\lambda \left( \mathbb{E}_{s^{(i)} \sim p(s^{(i)})} \left[ \frac{1}{\delta E\left( \pi_{\theta^{(i)}}\left(s^{(i)}\right) \right)} \right] - \frac{1}{\epsilon} \right) \right]_+,
    \end{equation}
    
\end{enumerate}
Here, $[\cdot]_+$ denotes projection onto the non-negative reals, and $\alpha_\theta$, $\alpha_\lambda$ are respective primal and dual learning rates controlling the speed of adjustments. 

\begin{remark}[Dual problem's behavior]
    
The Lagrange multiplier \( \lambda\left(s^{(i)}\right) \) is updated dynamically to reflect the constraint violation during the training at each term step for the $i$-th agent. For this, one can rewrite the update~\eqref{eq:dual_ex} as: 
\begin{equation}
\label{eq:dual}
\lambda\left(s^{(i)}\right) \gets \max\left(0, \lambda\left(s^{(i)}\right) - \alpha_\lambda \cdot \Tilde{\zeta}\left(a^{(i)}\right)- \beta \cdot \lambda\left(s^{(i)}\right)\right),
\end{equation}
with $\Tilde{\zeta}\left(a^{(i)}\right) = \left(\frac{1}{\delta E} - \frac{1}{\epsilon}\right  )$  being the constraint violation
and \( \beta \) is defined as a regularization parameter to penalize large values of \( \lambda\left(s^{(i)}\right) \), preventing unbounded growth. Therefore, our algorithm converges for any initial values of $\lambda\left(s^{(i)}\right) $ due to the convexity of the dual problem~\eqref{eq:dual}. We provide more details in \ref{app1} for~\eqref{eq:dual}, but generally it holds whether we use~\eqref{eq:dual} or~\eqref{eq:dual_ex}, as they are equivalent formulations. 

\end{remark}

Moreover, under suitable conditions (e.g., smoothness and boundedness of rewards and constraint costs), primal-dual methods converge to a saddle point $(\pi^\star, \lambda^\star)$ satisfying the Karush-Kuhn-Tucker (KKT) conditions for constrained optimization~\cite{boyd2004convex}:

\begin{align}
& \nabla_\theta \mathcal{L}(\theta^*, s^*) = 0, \label{eq:kkt1}\\
& \mathbb{E}_{s \sim p(s)}\left[ \frac{1}{\delta E(\pi_{\theta^*}(s)) }\right] \leq \frac{1}{\epsilon},\label{eq:kkt2} \\
& \lambda^* \geq 0,\label{eq:kkt3} \\
& \lambda^* \left( \mathbb{E}_{s \sim p(s)}\left[  \frac{1}{\delta E(\pi_{\theta^*}(s)) } \right] - \frac{1}{\epsilon} \right) = 0,\label{eq:kkt4}
\end{align}
The condition \eqref{eq:kkt1} is the direct consequence of the Lagrange definition~\eqref{eq:lagrange}. Updating the multiplier using \eqref{eq:dual_ex} implies the boundedness of the violation \eqref{eq:kkt2} and its positivity \eqref{eq:kkt3}. 
Thus, the learned policy $\pi^\star$ satisfies the reward maximization goal while respecting the constraint $\delta E(a) \leq \epsilon$ in expectation. The complementary slackness condition~\eqref{eq:kkt4} requires that either the constraint is inactive (slack), or the Lagrange multiplier is zero. This condition is satisfied during the dual step. 

\subsection{Constraint-Augmented Bellman Operator}

In RL, the Bellman operator plays a fundamental role by recursively defining the value function. In the standard setting, the Bellman operator $\mathcal{T}$ for a value function $V$ is given by~\cite{mousavi2018deep}
\begin{equation}
    (\mathcal{T}V)(s^{(i)}) = \max_{a^{(i)}} \left\{ R\left(s^{(i)},a^{(i)}\right) + \gamma \mathbb{E}_{s' \sim p(\cdot|s^{(i)},a^{(i)})} [V(s')] \right\},
\end{equation}

In RL-DAUNCE, we introduce a constraint-augmented Bellman operator $\mathcal{T}^\lambda$, defined as
\begin{equation}
\label{eq:bell}
(\mathcal{T}^\lambda V)(s{(i)}) = \max_{a{(i)}} \left\{ 
R\left(s^{(i)},a^{(i)}\right) - \lambda\left(s^{(i)}\right)  \cdot \left(\frac{1}{\delta E(a^{(i)})}- \frac{1}{\epsilon}\right)
+ \gamma \mathbb{E}_{s' \sim p(\cdot|s^{(i)},a^{(i)})} [V(s')]  \right\},
\end{equation}
that allows imposing soft constraints using penalized constraint violations.

Thus, actions that violate the physical constraint are explicitly penalized during both policy evaluation and improvement steps. Furthermore, we have $\gamma \in (0,1)$ and dual update~\eqref{eq:dual_ex} is independent of 
$V$ and does not affect the contraction property, since it appears identically for any arbitrary $V$. Therefore, similar to  $\mathcal{T}$~\cite{bellemare2016increasing}, $\mathcal{T}^\lambda$ remains a $\gamma$-contraction in the sup-norm:
\begin{equation}
    \|\mathcal{T}^\lambda V_1 - \mathcal{T}^\lambda V_2\|_\infty \leq \gamma \|V_1 - V_2\|_\infty,
\end{equation}
    ensuring convergence of iterative methods. 
 \eqref{eq:bell} explains the parameter $\lambda$ controls the balance between the constraint and reward maximization. As $\lambda$ increases, the optimal policy places more emphasis on constraint satisfaction relative to reward maximization. In the limit $\lambda \to \infty$, the agent fully prioritizes constraint satisfaction. Next, we need to reformulate the RL framework as primal-dual optimization using~\eqref{eq:lag}.

\begin{remark}
   The advantages of our dual gradient method are: automatic tuning of constraint enforcement strength~\cite{paternain2019learning}, balancing exploration and feasibility during learning, and scalability to multiple constraints by introducing a separate dual variable for each one \cite{paternain2022safe}.
\end{remark}

\subsection{RL-DA with constraints}
Finally, with all the definitions in this section, we introduce RL-DAUNCE framework that conserves physical constraints. For this, we incorporate
the updated Lagrange multiplier into the reward function to penalize constraint violations as:
\begin{equation}
\label{eq:reward_PD}
  R_{PD}(s,a) = R(s,a) - \lambda(s) \cdot \Tilde{\zeta}(a).
\end{equation}
Maximizing the primal-dual reward function $R_{PD}(s,a)$ ensures that the constraint \( \delta E(a) \leq \epsilon \) is enforced while dynamically adapting the Lagrange multiplier \( \lambda(s) \) at each iteration step. The dual updates are guided by the degree to which the constraint is satisfied or violated, allowing \( \lambda(s) \) to adjust by the problem's structure. This adaptive mechanism promotes convergence while maintaining stability and preventing excessive penalization.

\begin{remark}
    The constraint acts as an effective reward shaping term. It modifies the effective reward landscape by subtracting a penalty proportional to the violation magnitude. Consequently, the agent is guided to favor actions that simultaneously maximize task reward and respect the physical laws encoded by $\delta E(a)$. In particular, the standard Bellman operator applied to $R_{PD}$ in~\eqref{eq:reward_PD} recovers the constraint-augmented dynamics:
\begin{equation}
(\mathcal{T}^\lambda V)(s) = \max_{a} \left\{ R_{PD}(s,a) + \gamma \mathbb{E}_{s' \sim p(\cdot|s,a)} [V(s')] \right\}.
\end{equation}

Thus, enforcing constraints can be viewed equivalently as modifying the underlying environment reward structure, encouraging the emergence of physically-consistent behavior through learning.
\end{remark}

\begin{figure}
    \centering
    \hspace*{-1cm}\includegraphics[width=.9\linewidth]{ 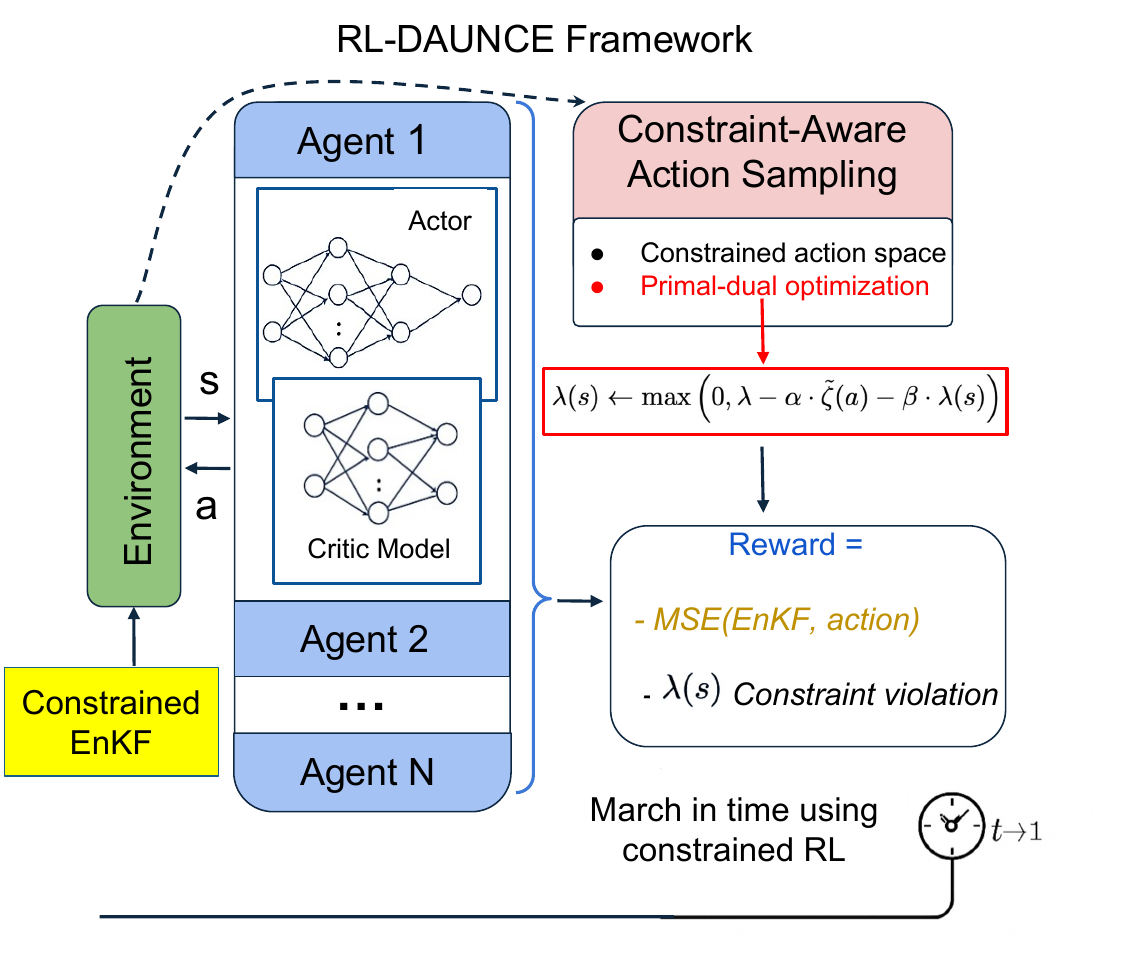}
    \caption{Overview of the proposed RL framework for constrained data assimilation. The RL agent ensemble learns to propose physically consistent actions based on EnKF-generated data, with constraints applied via primal-dual optimization during training. The system evolves sequentially in time with learned actions subject to positivity, conservation, and other constraints.}
    \label{fig:overview}
\end{figure}

To impose hard constraints directly on variable states at the next time step (e.g., lower $a_{min}$ or upper $a_{max}$ bounds on any of the estimated states), we define a constrained action space $ \mathcal{A}_c$ in which these bounds are enforced strictly. That is, the resulting state is guaranteed to remain above the lower bound and below the upper bound, ensuring the constraints are satisfied with absolute enforcement at each time step.

\begin{equation}
    \mathcal{A}_c = \left\{ a \in \mathbb{R}^n \ \big| \ a_{min} \leq a_i \leq a_{max}\ \text{for all} \ i \right\}.
\end{equation}
Consequently, the optimization problem becomes:
\begin{equation}
    \max_{\theta} \ \mathbb{E}_{s \sim p(s)} \left[ R_{PD}(s, \pi_\theta(s)) \right] \quad \text{subject to} \quad \pi_\theta(s) \in \mathcal{A}_c \quad \forall s \in \mathcal{S}.
\end{equation}
This is equivalent to restricting the policy to operate over a feasible subspace $\mathcal{M} \subset \mathbb{R}^n$, where $\mathcal{M} = \mathcal{A}_c$. Such a formulation is conceptually related to manifold optimization~\cite{hu2020brief}, where learning occurs over a constraint surface embedded in the ambient space. In practice, this is achieved by explicitly restricting the action space to $\mathcal{A}_c$ during training and inference. This design allows the RL agent to respect hard constraints by construction, without requiring external correction or projection steps. We provide an overview of our method in Figure~\ref{fig:overview}.
Once the ensemble of agents $\{\pi_\theta^{(i)}\}$ is trained, we perform inference by sampling from the ensemble policies, in analogy with state sampling in the EnKF. This not only maintains consistency with the EnKF’s philosophy of ensemble forecasting but also provides a principled approach for uncertainty quantification in the RL-based filter.

\begin{remark}[On enforcing hard constraints on state variables]
A common approach to enforce constraints, e.g., action positivity, is to apply a suitable activation function at the output layer. The technique relies on applying transformations such as \texttt{softplus}, \texttt{sigmoid}, or the exponential function to ensure that the output remains in a desired range (e.g., positive values). While effective for ensuring lower bounds (e.g., positivity), this method is generally less flexible for enforcing upper bounds. For example, \texttt{softplus} ensures outputs in \( (0, \infty) \), but cannot enforce finite upper limits without additional rescaling. Moreover, these transformations reshape the loss landscape and may introduce vanishing gradient issues, especially for saturating activations like \texttt{sigmoid}. Besides, uncertainty quantification is a challenge for this approach since the activation function appears after sampling. This can truncate or distort the action distribution near the boundaries, making it difficult to accurately estimate variance or confidence intervals. Overall, explicit action space constraints offer more general and precise control over the allowable actions.
\end{remark}

 \begin{remark}
     In data assimilation, a common but less principled approach to enforcing constraints such as positivity is to first estimate the state variable and then manually project or relax it post hoc (e.g., setting negative values to a small positive constant like $10^{-5}$. While simple, this technique can lead to inconsistencies between the learned policy and the true dynamics of the system, as the optimization process is unaware of the constraints during training. In contrast, constraining the action space directly ensures that constraint satisfaction is built into the policy through constraint-aware optimization. This results in more coherent behavior, smoother training, and better generalization.
 \end{remark}


\section{Applying RL-DAUNCE to an Intermittent Atmospheric System}
\label{sec:MJO}

In this section, we demonstrate the skill of RL-DAUNCE in estimating the state variables of the Madden-Julian Oscillation (MJO). The study is based on a perfect model setup: the observations are given by a single realization of the stochastic skeleton model of the MJO. This model captures the complex temporal and spatial evolution of the MJO, including its challenging intermittent and nonlinear behaviors.

\subsection{Overview of the MJO and the stochastic skeleton model}

The MJO is the dominant mode of tropical intraseasonal variability, characterized by a slow-moving planetary-scale envelope of convection that propagates eastward across the equatorial Indian and western/central Pacific Oceans \cite{zhang2005madden, zhang2020four}. The MJO influences both tropical and extra tropical weather patterns and plays a crucial role in modulating large-scale climate phenomena such as the El Ni\~no-Southern Oscillation.

The MJO skeleton model describes the large-scale dynamics of the MJO. It captures several key features of the MJO, including (a) a slow eastward propagation speed ($\sim5$ m/s), (b) a peculiar dispersion relation with near-zero group velocity ($d\omega/dk \approx 0$), (c) a quadrupole structure in the large-scale circulation, (d) the intermittent generation of MJO events, and (e) the organization of MJO events into wave trains exhibiting growth and decay.
The MJO skeleton model is derived from the three-dimensional primitive equations for the zonal velocity $u$, meridional velocity $v$, and potential temperate $\theta$. Following standard practice, we project these equations onto the first baroclinic mode in the vertical dimension:
\begin{equation}
\begin{aligned}
   \frac{\partial u}{ \partial t} - y v - \frac{\partial \theta}{ \partial x} &= 0, \label{eq:u_eq} \\
    y u - \frac{\partial \theta}{ \partial y} &= 0,  \\
    \frac{\partial \theta}{ \partial t} -  \frac{\partial u}{ \partial x} -  \frac{\partial v}{ \partial y} &= \bar{H} a - s^{\theta}, \\
    \frac{\partial q}{ \partial t} + \tilde{Q} \left( \frac{\partial u}{ \partial x} -  \frac{\partial v}{ \partial y}\right) &= - \bar{H} a + s^q, \\
     \frac{\partial a}{ \partial t} &= \Gamma aq + \sqrt{\Delta a \Gamma |q| a} \, \dot{W}(t),
\end{aligned}
\end{equation}
 where the last equation is the continuous approximation of the birth–death process~\cite{chen2016filtering}. Next, we project onto the first meridional mode as 
 \begin{equation}
    a(x,y,t) = \left(\bar{A}(x) + A(x,t)\right) \Phi_0,
\end{equation}
 with $\Phi_0 = \pi^{-1/4} \exp(-y^2/2),$. This projection yields a simplified model that depends exclusively on the zonal coordinate $x$ along the equator and time $t$. The model reads \cite{thual2014stochastic, majda2009skeleton, majda2011nonlinear, chen2016filtering}:
\begin{align}
    \frac{\partial K}{\partial t} + \frac{\partial K}{\partial x} &= \frac{1}{2} \left( S^\theta - \overline{H}A\right),     \label{eq:K}\\
    \frac{\partial R}{\partial t} - \frac{1}{3} \frac{\partial R}{\partial x} &= \frac{1}{3} \left( S^\theta - \overline{H}A \right), \\
    \frac{\partial Q}{\partial t} + \tilde{Q} \left( \frac{\partial K}{\partial x} - \frac{1}{3} \frac{\partial R}{\partial x} \right) &= \left( \frac{\tilde{Q}}{6} - 1 \right) \left( \overline{H}A - S^q \right), \\
    \frac{\partial A}{\partial t} &= \Gamma \left( A+\bar{A} \right)Q + \sqrt{\Gamma |Q| \left( A+\bar{A} \right)} \, \dot{W}(t),
    \label{eq:trunc}
\end{align}
Here, the variables $K$, $R$, $Q$, and $A$ denote Kelvin wave, Rossby wave, moisture, and the envelope of the convective activity. The two physical variables $u$ and $\theta$ in the projected space are linear combinations of the two wave variables $K$ and $R$ (see \cite{thual2014stochastic} for details). The constants $\tilde{Q}$ and $\overline{H}$ are the mean background moisture and scaling constant for convective activity, respectively. In addition, $S^\theta$ and $S^q$ are projected source terms for temperature and moisture, $\Gamma$ is the growth rate coefficient, and $W$ is a white noise. Notably, all model variables, except $A$, are anomalies. The nonlinear interactions between $Q$ and $A$ form a nonlinear oscillator, while the state-dependent noise in the equation of $A$ triggers intermittency and extreme events. The stochastic skeleton model has been shown to capture the observed propagation and the statistics of the MJO \cite{stachnik2015evaluating, ogrosky2015mjo}. It provides a useful framework for studying the MJO's fundamental mechanisms, particularly in the context of data assimilation and prediction.

\subsection{Sources of constraints: positivity of convective activity and energy conservation}
The first constraint in the MJO skeleton model is the positivity of the convective activity $A$ satisfying $A+\bar{A} > 0$, which is due to its physical representation.

The second feature of the MJO skeleton model regards its energy. In the absence of stochastic forcing and balanced source terms, namely \( s^\theta(x) = s^q(x):=s \), the model conserves a positive total energy~\cite{stechmann2015identifying}:
\begin{equation}
\partial_t \left(
\frac{1}{2} u^2 +
\frac{1}{2} \theta^2 +
\frac{1}{2} \tilde{Q} (1 - \tilde{Q}) \left( \frac{\theta + q}{\tilde{Q}} \right)^2 +
\frac{\bar{H}}{\Gamma \tilde{Q}} a - \frac{s}{\Gamma \tilde{Q}} \log a
\right)
- \partial_x (u \theta) - \partial_y (v \theta) = 0.
\label{eq:total_energy}
\end{equation}
where $q$ and $a$ are the moisture and convective activity before applying the meridional truncation.
This total energy is a sum of four terms: dry kinetic energy \( \frac{1}{2} u^2 \), dry potential energy \( \frac{1}{2} \theta^2 \), a moist potential energy proportional to \( \tilde{Q}^{-1} (\theta + q)^2 \) (see, e.g., \cite{frierson2004large}), and a convective energy \( \frac{\bar{H}}{\Gamma \tilde{Q}} a - \frac{s}{\Gamma \tilde{Q}} \log a \). Note that the natural requirement on the background moisture gradient, \( 0 < \tilde{Q} < 1 \), is needed to guarantee a positive energy. In addition, this energy is a convex function of \( u, \theta, q, \) and \( a \). The meridionally truncated system in \eqref{eq:K}--\eqref{eq:trunc} possesses similar conservation laws~\cite{chen2015nonlinear}:
\begin{equation}
\label{eq:energy}
E =  K^2 + \frac{3}{8} R^2 + \frac{1}{2} \frac{\tilde{Q}}{1 - \tilde{Q}}
\left( \frac{Q}{\tilde{Q}} - {K}- {R} \right)^2
+  \frac{S}{\Gamma \tilde{Q}} \left( \left( A+\bar{A} \right)- \log \left( A+\bar{A} \right) \right).
\end{equation}
Notably, when stochastic noise appears, total energy is not strictly conserved. This is also consistent with nature, where small-scale perturbations and damping effects modify the energy to a certain extent. Nevertheless, the total energy fluctuates around a certain level. Imposing such a soft constraint can still play a crucial role in improving data assimilation skills.
\subsection{Setup of the EnKF and the RL-DAUNCE}
The stochastic and nonlinear nature of the MJO makes it a compelling test case for data assimilation. Conserving the energy and positivity of $A$ during the process is crucial to obtaining accurate results. However, this is highly nontrivial. For example, manually forcing $A$ to positive values as post-processing results in unstable solutions~\cite{gleiter2022ensemble}. The RL-DAUNCE framework enhances the ability to estimate the state of the MJO while maintaining physically consistent solutions.

For the numerical simulation, we discrete the spatial domain with \( N_a = 64 \) grid points, corresponding to an equatorial length of 40{,}000 km and a dimensionless zonal length unit of 15{,}000 km, resulting in a total nondimensional domain length of \( L = 8/3 \). The grid spacing was set as \( dx = L/N_a \), and the time step was fixed at \( dt = 0.001 \), which corresponds to approximately \(1.44\) hours under the time unit of two months. The convective activity growth rate set as \( \Gamma = 1.66 \), and the climatological background moisture \( \tilde{Q} = 0.9 \). Also, we set \( \bar{H} = 0.22 \), with source terms \( S^q = S^\theta = \bar{A} \bar{H}=0.022 \), which is known as the spatially homogeneous case. In the second example, we model background radiative cooling, background latent heating ($S^q,\,S^\theta$) as warm-pool following \cite{majda2019tropical}:
\begin{equation}\label{Warm_Pool_Background}
    S^q_{WP}= S^\theta_{WP} = 0.022 \times \left(1-0.6 \cos \left( \frac{2\pi x}{L}\right)\right).
\end{equation}

\begin{figure}[h]
    \centering
    \includegraphics[width=0.95\linewidth]{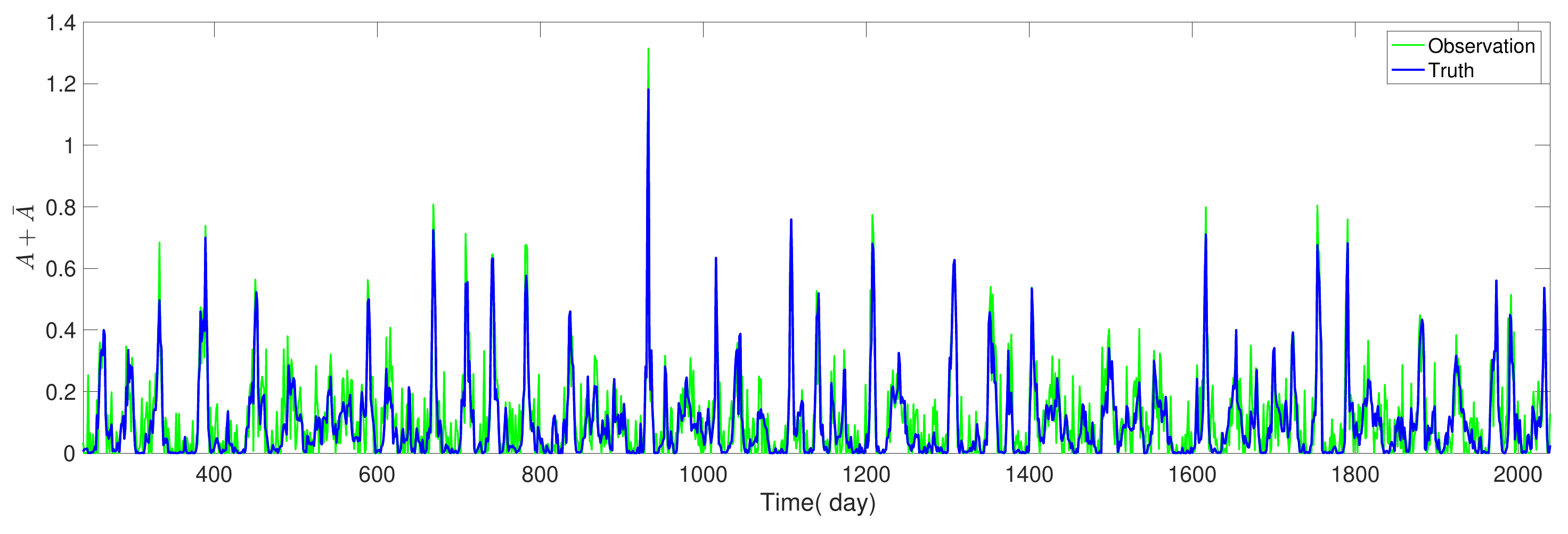}
    \caption{Temporal evaluation of variable $a$ and its corresponding observation at a fixed grid point.}
    \label{fig:obs_a}
\end{figure}

The observational variables in filtering the stochastic skeleton model involve only the convective activity $a$. This situation is consistent with investigating and operational monitoring of the real-world MJO~\cite{szekely2016extraction}. Since $A+\bar{A}$ is positive, we add noise using a lognormal distribution with zero mean and variance $0.0063$, which is 7.5\% of the variance of $A+\bar{A}$. The observations are recorded every $28.8$ hours. We observe $A$ in all spatial grid points. Figure~\ref{fig:obs_a} shows the truth variable $A+\bar{A}$ and its corresponding observation at a fixed spatial location. Clearly, the signal of $A$ is highly intermittent. The PDF of $A$ is highly non-Gaussian with a one-side fat tail corresponding to extreme events.

In generating the EnKF solution, we use an EAKF solver \cite{anderson2001ensemble} and apply the Gaspari-Cohn localization matrix, which filters out weak correlations beyond a certain distance. Given that our system consists of 64 spatial points, we element-wise multiply this localization matrix with the variables at their respective spatial points. Only the nonzero localized values are provided as inputs to the RL model, ensuring that the model learns from spatially relevant information while reducing computational complexity.

To train our RL approach, we define the state as the set of variables \((K, R, Z, A)\), with $Z$ being a variable delivering $Q= Z + \tilde{Q} (K + R)$. The action corresponds to predicting the values of these variables at the next time step \(t + \delta t\), using the observations of these variables at the same spatial point and at the current time \(t\). Additionally, we incorporate the temporal variations of these variables by computing their finite differences at two previous time steps, normalized by the time step \(\delta t\). The time \(t + \delta t\) and the spatial point at which we aim to estimate the variables are also included in the state representation.
In addition, we normalized the input observations for RL such that each state variable remains within $[-1,1]$ to ensure stable learning.


To ensure that the energy variation remains within acceptable bounds, we impose a constraint on the total energy of the predicted system. Specifically, we define a tolerance interval $\epsilon = [0.015, 0.08]$, representing the permissible range for energy deviations. This aligns with physics, as the damping and random forcing alter the total energy over time, though the energy remains approximately stable around a certain level. At each training step during the RL process, we update the Lagrange multiplier $\lambda$ to penalize deviations of the total energy of the RL's predicted actions from this specified interval. Each agent initializes $\lambda$ with a value of one. This dynamic adjustment of $\lambda$ throughout training enables the framework to effectively enforce the energy constraint, thereby estimating the system's variable states accurately. 

Furthermore, it is essential to enforce positivity preservation of $A+\bar{A}$ as a hard constraint, given its critical importance for the stability of the estimation process and the physical interpretability of convection activity. To achieve this, the action space corresponding to the state variable 
$A$ is bounded from below by $-\bar{A}$.

\subsection{Evaluation scores}
To evaluate the performance of the constrained EnKF and the RL-DAUNCE in recovering the large-scale MJO features through error quantities. Standard skill scores, i.e., the root-mean-squared-error (RMSE) and pattern correlation (Corr), are adopted. Note that the RMSE is normalized by the standard deviation of the true MJO field and averaged over all spatial grid points. At each time instant, the RMSE is calculated as:
\begin{equation}
\text{RMSE}(t) = \frac{ \sqrt{ \frac{1}{s} \sum_{i=1}^{s} \left( MJO_i^{\text{truth}}(t) - MJO_i^{\text{est}}(t) \right)^2 } }{ \sigma_{\text{truth}}(t) },
\end{equation}
where \( s \) is the number of spatial points, \( u_i^{\text{truth}}(t) \) is the true value at grid point \( i \) and time \( t \), \( u_i^{\text{est}}(t) \) is the estimated value (from EnKF or RL-DAUNCE), and \( \sigma_{\text{truth}}(t) \) is the standard deviation of the true MJO field at time \( t \).
The Corr is defined as:
\begin{equation}
\text{Corr}(t) = \frac{ \text{Cov} \left( MJO^{\text{truth}}(t), MJO^{\text{est}}(t) \right) }{ \sigma_{\text{truth}}(t) \sigma_{\text{est}}(t) },
\end{equation}
where \( \text{Cov}(\cdot, \cdot) \) denotes the covariance, and \( \sigma_{\text{est}}(t) \) is the standard deviation of the estimated MJO field at time \( t \).

\subsection{DA Results}
Table~\ref{tab:comparison} summarizes the RMSE and Corr values at four different time points. Overall, the RL method exhibits close  RMSE and Corr to the constrained EnKF, indicating similar reconstruction of the MJO large-scale features.

\begin{figure}[h]
    \begin{flushleft}
    \hspace*{-1.7cm}
\includegraphics[width=1.2\linewidth]{ 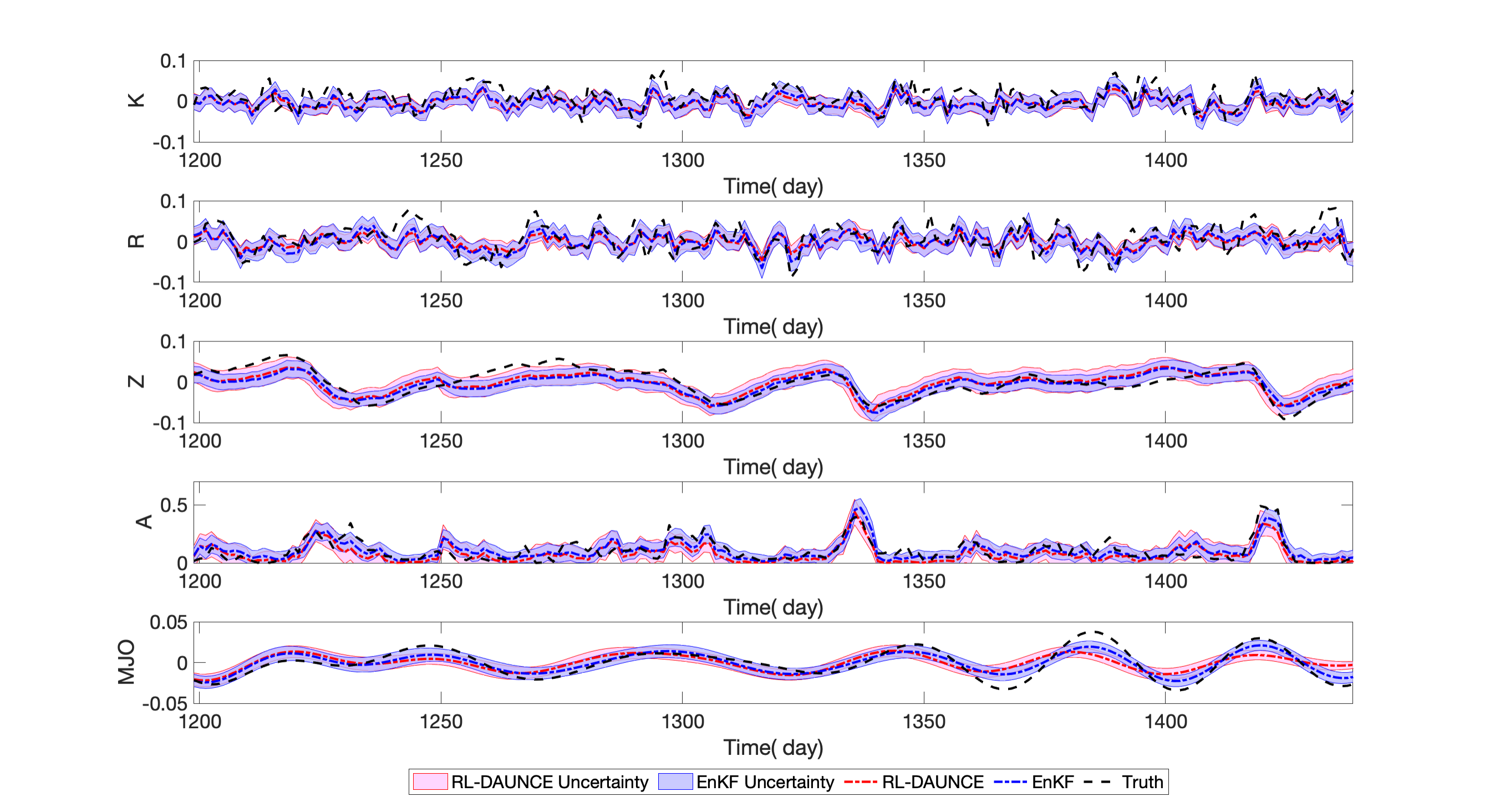}
        \end{flushleft}
    \caption{Temporal trajectories of the state variables $K, R, Z, A$, and MJO at a fixed spatial location. Each subplot compares the ground truth (black), the mean and uncertainty ($\ pm2$ standard deviation) from the constrained EnKF (blue), and those from the RL-DAUNCE framework (red). The RL-DAUNCE predictions closely follow the constrained EnKF in both the mean state and uncertainty, demonstrating the RL-DAUNCE's ability to replicate EnKF-like assimilation performance.}
    \label{fig:uncertain}
\end{figure}

In Figure~\ref{fig:uncertain}, we compare the results of the constrained EnKF and RL-DAUNCE in terms of both the mean state and the uncertainty, against the true state (ground truth). The shaded regions represent the uncertainty based on the ensemble spread around the mean. Notably, the uncertainty region produced by the RL-DAUNCE closely matches that of the constrained EnKF, indicating that RL-DAUNCE rigorously captures the underlying uncertainty structure. Furthermore, the mean state trajectory of all variables estimated by RL-DAUNCE remains consistent with that of the constrained EnKF. It follows the true state closely, even though the observable only involves the noisy signal of $A$. In particular, RL-DAUNCE successfully recovers the intermittent extreme events appearing in the signal of the convective activity $A$. This demonstrates RL-DAUNCE's ability to learn both accurate state estimates and reliable uncertainty quantification, effectively replicating the behavior of the constrained EnKF. The accurate estimate of the state variables also guarantees the success in recovering the amplitude and propagation of the MJO, as will be seen below. As a remark, applying the standard unconstrained EnKF leads to a rapid blow-up of the solution due to violating physical constraints. In particular, instability arises when the estimated convective activity $A$ becomes negative, ultimately causing the solution to diverge. Furthermore, it has been shown in  \cite{gleiter2022ensemble} that violating the energy constraint can still result in significantly larger errors even when the solution does not blow up.

\begin{figure}[htbp]
  \centering

  \hfill
  \begin{subfigure}[b]{0.49\textwidth}
    \includegraphics[width=\textwidth]{ 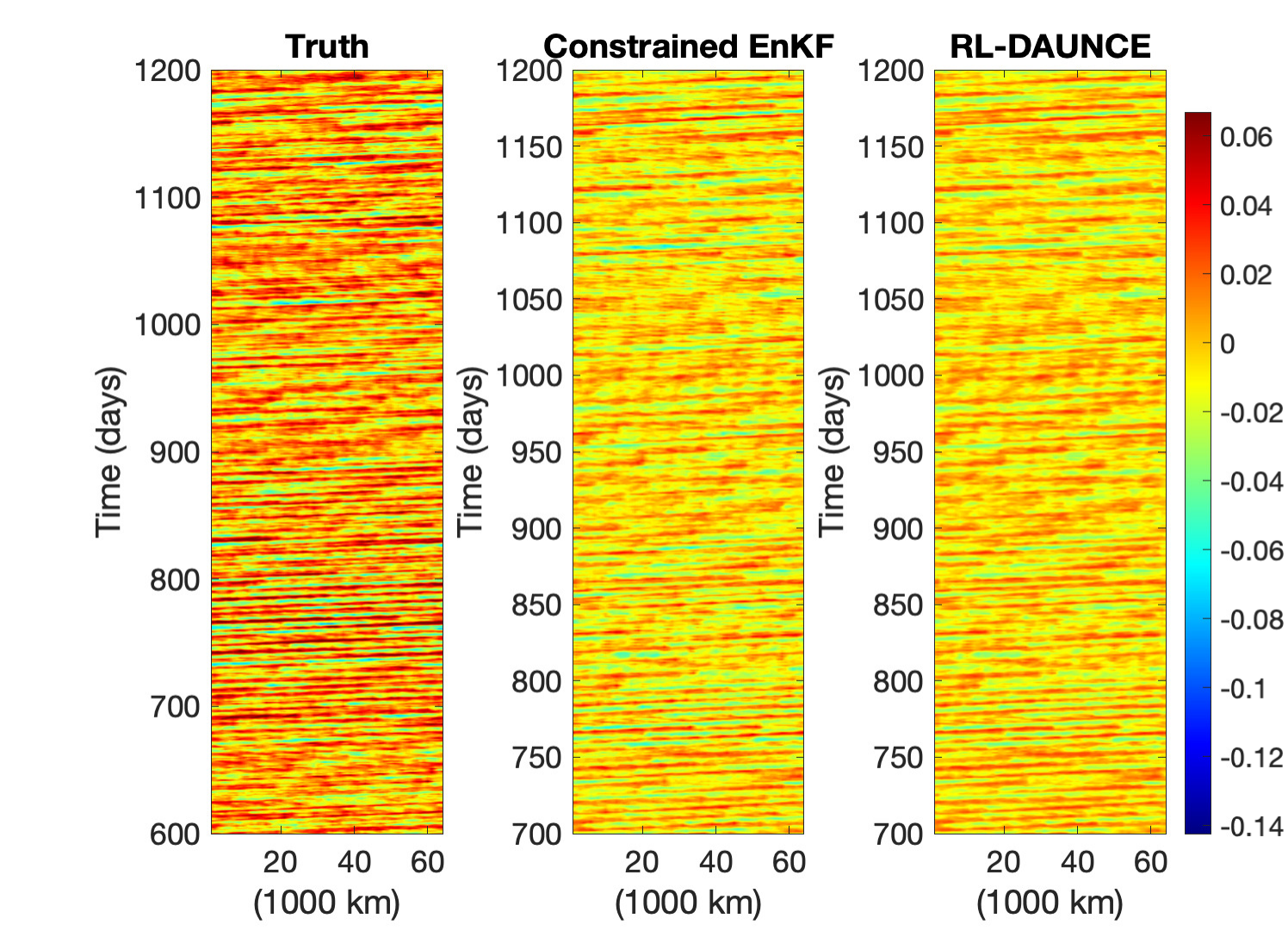}
    \caption{K}
  \end{subfigure}
  \begin{subfigure}[b]{0.49\textwidth}
    \includegraphics[width=\textwidth]{ 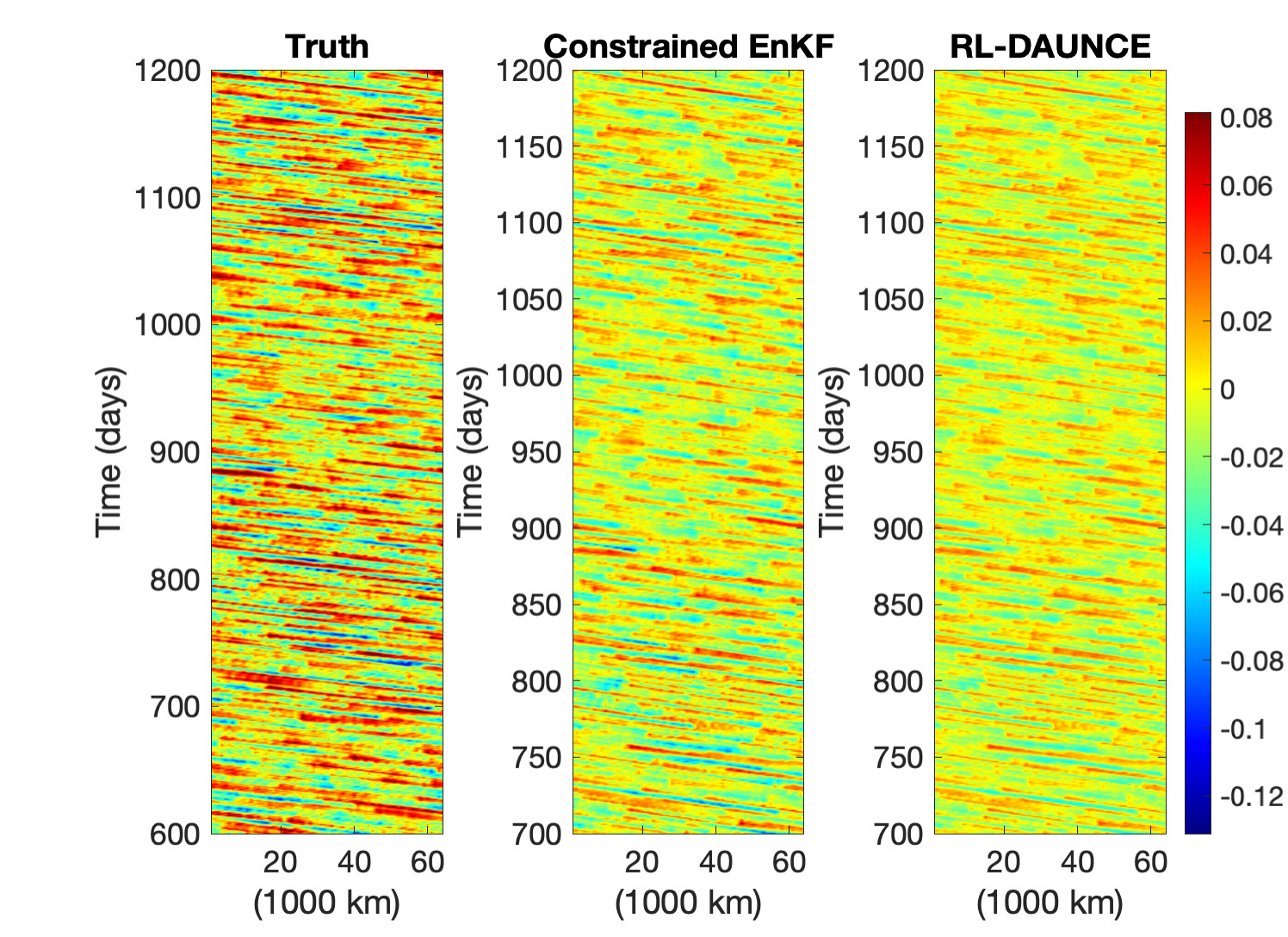}
    \caption{R}
  \end{subfigure}

  \hfill
  \begin{subfigure}[b]{0.49\textwidth}
    \includegraphics[width=\textwidth]{ 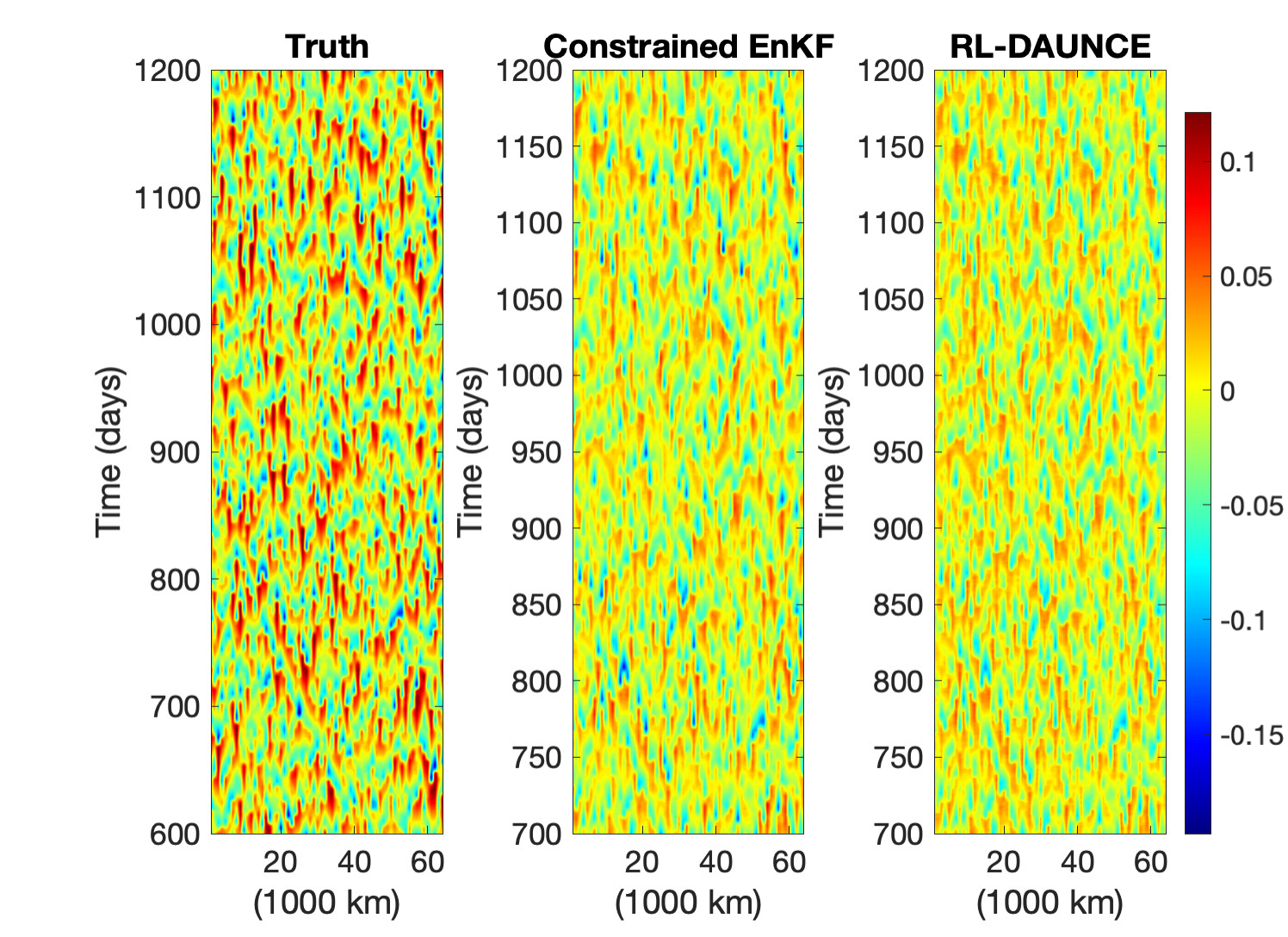}
    \caption{Z}
  \end{subfigure}
  \begin{subfigure}[b]{0.49\textwidth}
    \includegraphics[width=\textwidth]{ 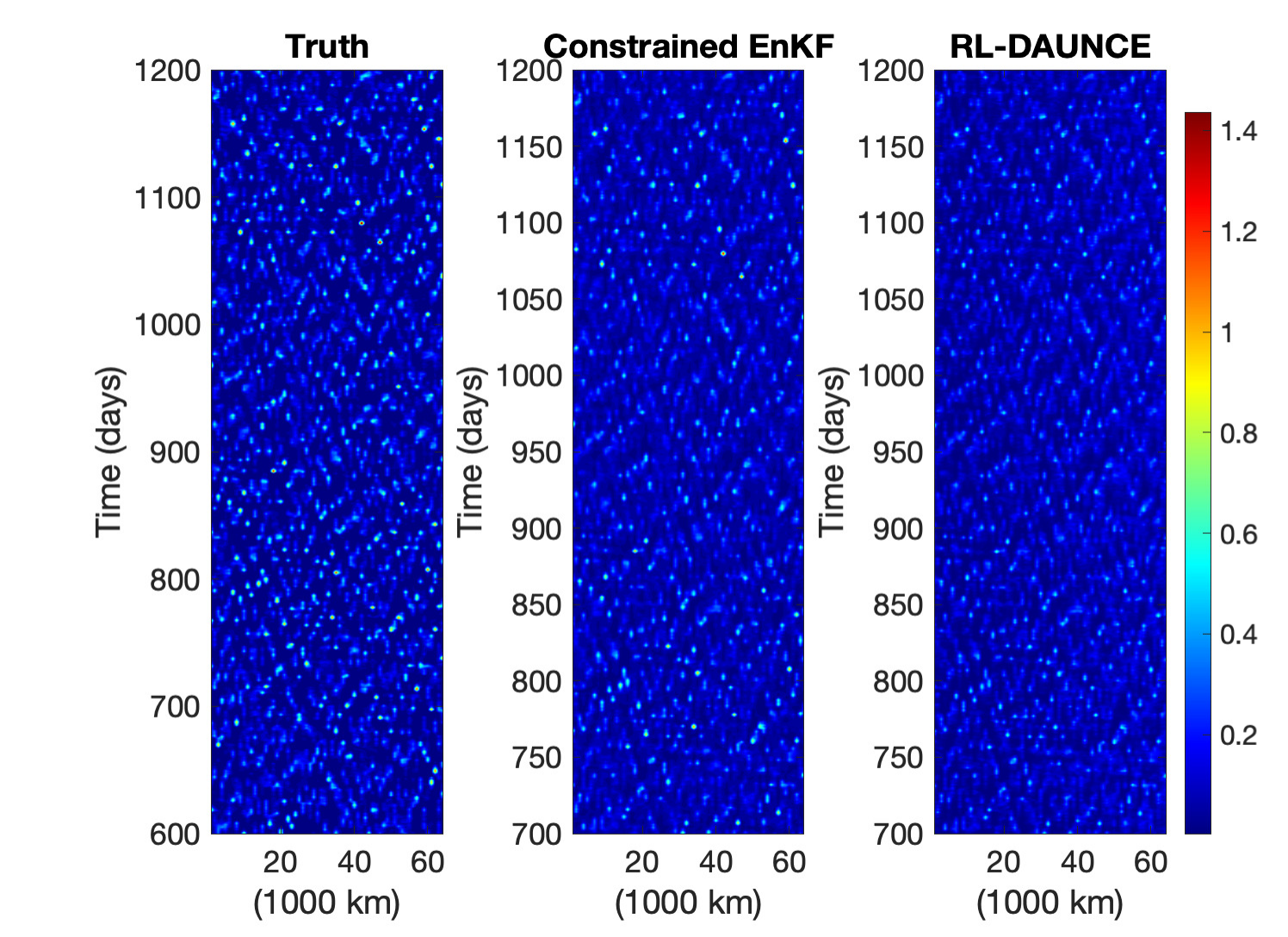}
    \caption{A}
    \label{fig:hovmoller_diff_K}
  \end{subfigure}

  \begin{subfigure}[b]{0.6\textwidth}
    \includegraphics[width=\textwidth]{ 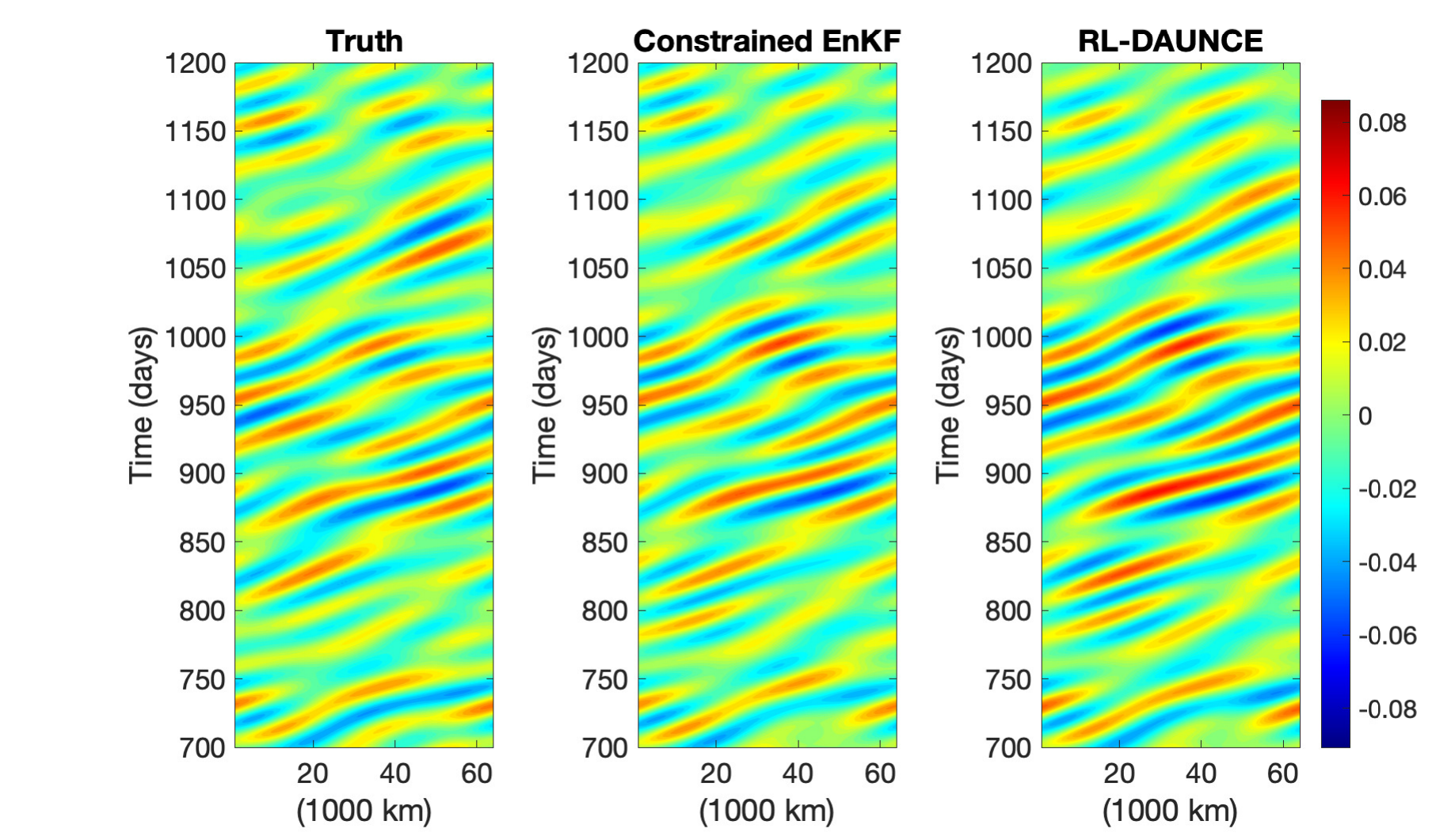}
    \caption{MJO}
  \end{subfigure}
\caption{Hovmöller diagrams of the state variables $K, R, Z, A$, and MJO in the space-time domain.}
       \label{fig:Hov}
\end{figure}

In Figure~\ref{fig:Hov}, we present Hovmöller diagrams for the state variables and MJO, visualizing their evolution across space and time. We compare each variable's spatiotemporal structures obtained from the ground truth, the constrained EnKF, and RL-DAUNCE. RL-DAUNCE reconstructions exhibit strong agreement with both the EnKF results and the true states, successfully capturing the dominant wave patterns, propagation characteristics, variability across the space-time domain, and intermittent phenomena. These results further highlight RL-DAUNCE's capacity to learn complex dynamical behaviors and serve as a computationally efficient approach for constrained data assimilation.

\begin{table}[h!]
\centering
\caption{Comparison of RMSE and Corr between constrained EnKF and RL-DAUNCE at different time points.}
\label{tab:comparison}
\begin{tabular}{c|cc|cc}
\hline
\textbf{Time (day))} & \textbf{RMSE (EnKF)} & \textbf{RMSE (RL-DAUNCE)} & \textbf{Corr (EnKF)} & \textbf{Corr (RL-DAUNCE)} \\
\hline
700 & 0.2483 & 0.3056 & 0.8726 & 0.8036 \\
800 & 0.1747 & 0.1817 & 0.9385 & 0.9296 \\
950 & 0.1028 & 0.1151 & 0.9835 & 0.9843 \\
1100 & 0.1901 & 0.2961 & 0.9427 & 0.8065 \\
\hline
\end{tabular}
\end{table}

\begin{figure}[h]
    \centering
    \includegraphics[width=1\linewidth]{ 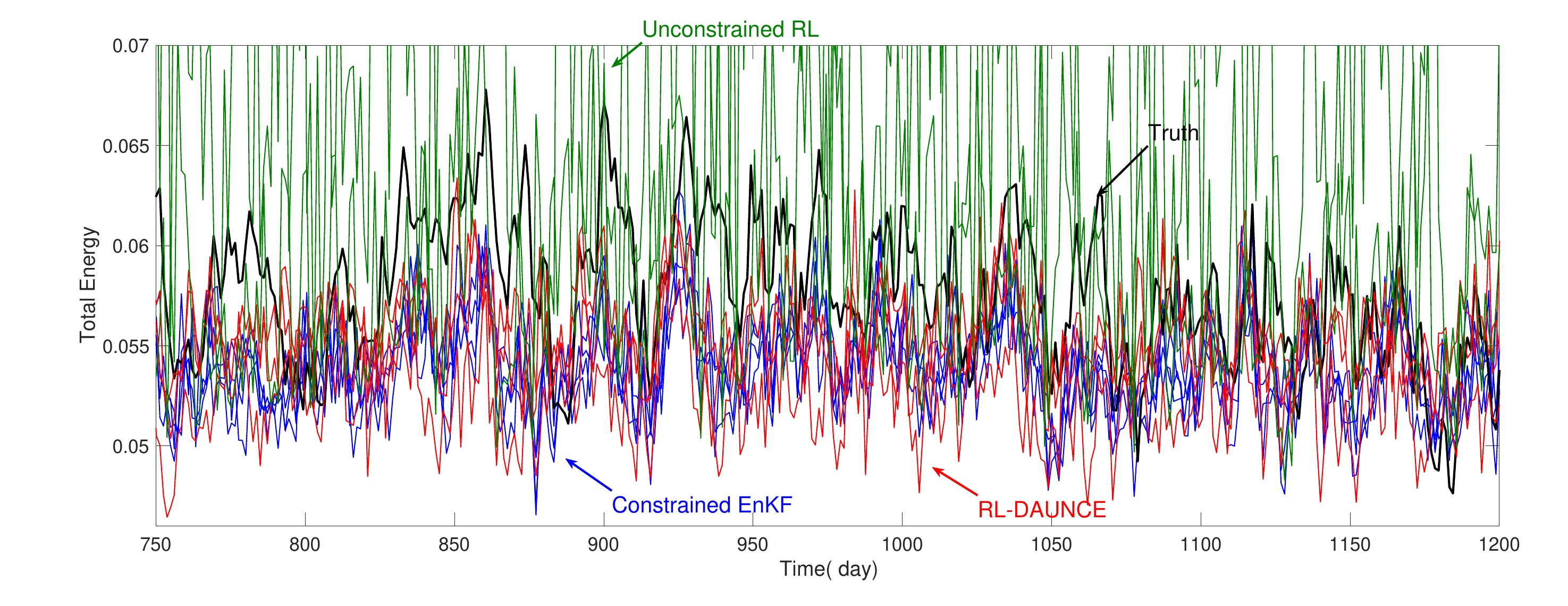}
    \caption{Comparison of the total energy across ensembles for different methods. RL-DAUNCE successfully conserves total energy through the deployment of our constraint enforcement algorithm. However, without applying the constraint enforcement, even though RL was trained using constrained EnKF data, the total energy is not conserved. This highlights the critical role of constraint enforcement in preserving physical properties. Each dashed line represents the energy evolution of one ensemble. }
    \label{fig:energy}
\end{figure}

Finally, in Figure~\ref{fig:energy}, we show how the total energy evolves within a specific range for each ensemble in our framework. We observe that when RL-DAUNCE is trained using data generated by the constrained EnKF (with soft constrained on the energy and hard constraints on the positivity preserving) and our constraint enforcement algorithm is applied during prediction, the total energy effectively remains in the interval across all ensembles. Moreover, the range of the fluctuations in the RL-DAUNCE and the constrained EnKF is comparable. In contrast, the total energy is no longer preserved if the constraint enforcement is omitted during the training of RL, even though the RL model was trained on the same ENKF data. This comparison highlights the essential role of constraint enforcement in ensuring that the learned model respects the underlying physical laws. Without explicit enforcement using RL-DAUNCE, accumulated errors can lead to a significant drift in energy over time, even if the training data itself satisfies conservation properties. As a result, the estimated state contains significant errors as time evolves. It should also be noted that without the positivity constraint on the convection activity, the estimations are not stable. 

In terms of computational efficiency, RL-DAUNCE significantly outperforms both constrained and unconstrained EnKF methods. Using wall-clock time measurements, the constrained EnKF requires approximately 22.96 seconds per assimilation step, primarily due to the computational overhead of solving optimization problems with nonlinear constraints. Even the unconstrained EnKF still requires 5.61 seconds per step despite eliminating constraint enforcement, and more importantly, often yields unphysical posterior estimates due to the absence of structural or physical constraints. In contrast, RL-DAUNCE achieves remarkable speed, completing each assimilation step in just 1.1 seconds, while maintaining physical consistency through learning from data generated by the constrained EnKF. This demonstrates that RL-DAUNCE not only retains the accuracy and physical integrity of constrained assimilation but does so with a 20x speed-up over constrained EnKF and 5x over unconstrained EnKF, offering a compelling solution for real-time or large-scale data assimilation tasks.

\begin{figure}[h]
    \centering
    \includegraphics[width=1\linewidth]{ 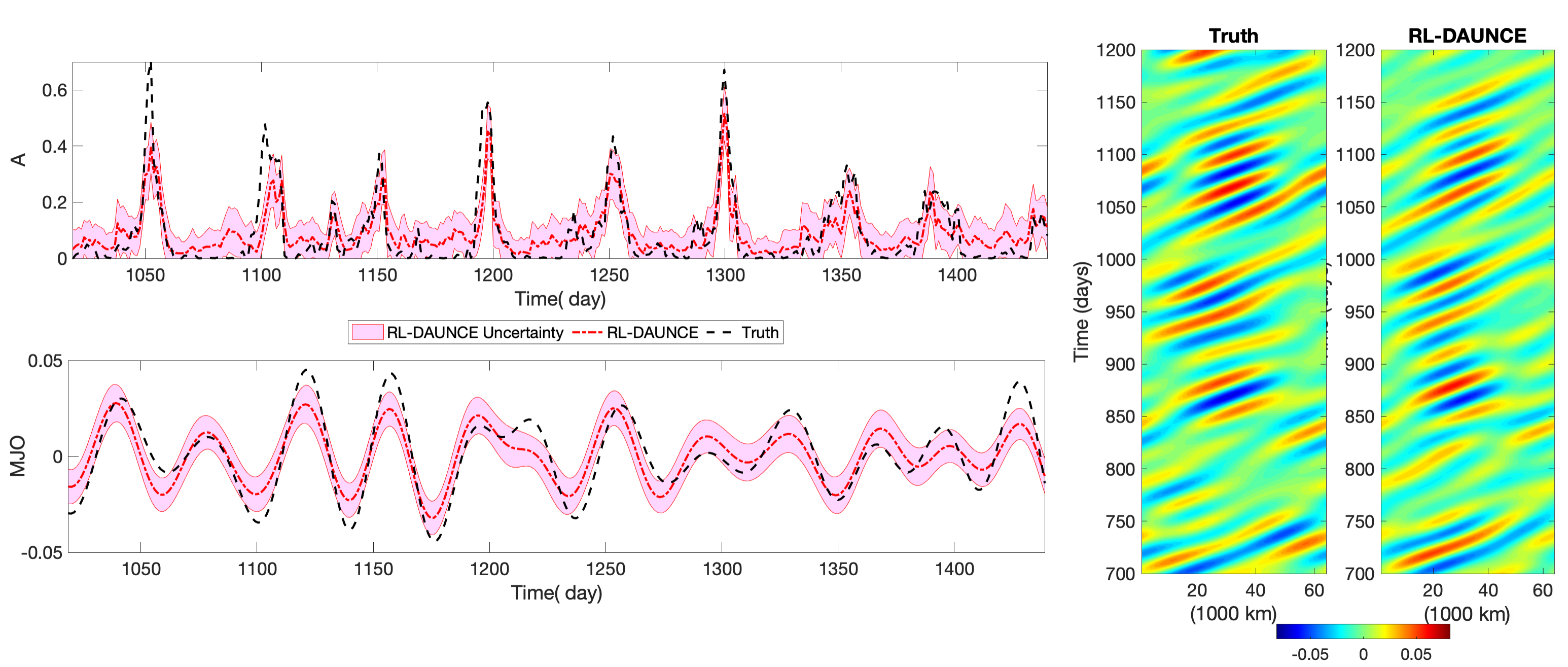}
    \caption{ Results for the warm pool case. Left: Time series of the state variables $A$ and MJO at a fixed spatial point, comparing RL-DAUNCE (red) with the ground truth (black). The shaded regions represent the uncertainty ($\pm 2$ standard deviation) estimated by RL-DAUNCE. Right: Hovmöller diagrams of the MJO variable from the RL-DAUNCE framework and the ground truth, showing the spatiotemporal propagation of MJO. The RL-DAUNCE results closely follow the true dynamics in both time-series behavior and spatial-temporal evolution.}
    \label{fig:warm}
\end{figure}

Finally, the setups for the warm-pool scenario remain the same as described above, except using the warm-pool heating and moisture profiles \eqref{Warm_Pool_Background}. The results are shown in Figure~\ref{fig:warm}, which demonstrates the same skill as the spatially homogeneous case. In particular, the intermittent extreme events in the convective activity are successfully recovered by RL-DAUNCE. In addition, the MJO strengths, phases, and wave propagations are all captured accurately.

\section{Conclusion}
\label{sec:conclude}
In this paper, we develop RL-DAUNCE (Reinforcement Learning-Driven Data Assimilation with Uncertainty-Aware Constrained Ensembles), a new RL-based method that enhances data assimilation with physical constraints. Fundamentally different from empirical post-hoc corrections, RL-DAUNCE enforces fundamental laws (e.g., energy conservation, positivity preservation) through two intrinsic mechanisms: (1) a primal-dual optimization strategy that dynamically penalizes constraint violations during training, and (2) hard bounds on the RL action space to preserve state variable validity. RL-DAUNCE has several unique features. First, it has an ensemble-inspired architecture, where RL agents mirror ensemble members, maintaining compatibility with conventional DA while leveraging the adaptive learning of the RL. Second, RL-DAUNCE facilitates the uncertainty quantification. This is achieved by letting each agent evolve independently. The ensemble statistics capture full distributional information beyond mean-state estimates. Third, RL-DAUNCE highlights the physically constrained assimilation. A primal-dual optimization scheme enforces constraints via dynamic reward penalties, while action-space bounds ensure state variable validity. These are all computationally efficient. RL-DAUNCE is robust and flexible enough to handle nonlinear, multivariable, and intermittent phenomena.

There are several promising future directions. First, further reductions in computational cost remain possible. Specifically, exploring transfer learning to adapt trained agents across similar dynamical systems could significantly lower training costs. Second, RL-DAUNCE shows potential for solving multi-model data assimilation problems and handling structural model uncertainty. Additionally, incorporating computational strategies from deep RL could allow RL-DAUNCE to address the high-dimensional problems encountered in real applications.

\section*{Acknowledgment}
The research of N.C. is funded by the Office of Naval Research N00014-24-1-2244 and the Army Research Office W911NF-23-1-0118. P. B. is partially supported as a research associate under the first grant. 
 
\bibliography{references_cleaned}

\appendix
\section{Dual Optimization and Gradient-Based Updates}
\label{app1}
It is important to emphasize that while primal-dual algorithms are designed to solve the dual formulation of a constrained optimization problem, they do not necessarily yield direct access to the primal solution. However, by establishing a bound on the duality gap (see e.g. \cite[Theorems 3 and 4]{paternain2022safe}), one can argue that solving the dual problem offers a sufficiently accurate approximation to the primal optimum.

A central mechanism used in our method is dual gradient descent. Let \( \nabla_\lambda d_\theta(\lambda) \) denote the gradient of the dual function \( d_\theta \). The update rule for dual gradient descent is expressed as:
\[
\lambda^{k+1} = \left[ \lambda^k - \alpha_\lambda \nabla_\lambda d_\theta(\lambda^k) \right]^+,
\]
where \( \alpha_\lambda > 0 \) is the step size, and \( [\cdot]^+ \) denotes projection onto the non-negative orthant of \( \mathbb{R}^m \). This projection step is crucial to ensure that the Lagrange multipliers \( \lambda \) remain non-negative, as required by the inequality constraints.

Under standard convexity and smoothness conditions, this algorithm is guaranteed to converge to a neighborhood of the optimal dual solution. The step size \( \alpha_\lambda \) controls the convergence rate and stability of the updates.

The gradient of the dual function is computed by evaluating the constraint function at the primal maximizer of the Lagrangian. More precisely, the gradient at iteration \( k \) is given by:
\[
\nabla_\lambda d_\theta(\lambda^k) = U(\theta^\star(\lambda^k)) - \frac{1}{\epsilon},
\]
where \( \theta^\star(\lambda^k) \) is the maximizer of the Lagrangian with respect to the primal variables at the current value \( \lambda^k \), \( U(\theta^\star(\lambda^k)) \) denotes the constraint evaluation (e.g., inverse of energy deviation \( (\delta E)^{-1} \)) at this optimal primal point, \( \epsilon \) is the user-defined threshold associated with the constraint.

This formulation forms the core of our dual optimization framework with dynamically adjusted Lagrange multipliers and enables robust constraint satisfaction throughout learning.

\end{document}